\documentclass[11pt,a4paper]{article}
\usepackage[hyperref]{acl2021}
\usepackage{times}
\usepackage{latexsym}

\usepackage{microtype}

\aclfinalcopy 

\usepackage{amsmath,amssymb,amsfonts}
\usepackage{algorithmic}
\usepackage{graphicx}
\usepackage{textcomp}
\usepackage{xcolor}
\usepackage{enumitem}
\usepackage{todonotes}
\usepackage{bbm}
\usepackage{amsmath}
\usepackage{xcolor}
\usepackage{subcaption}
\usepackage{graphicx}
\usepackage{multirow}
\usepackage{comment}
\usepackage{hyperref}
\usepackage{amsthm, amssymb}
\theoremstyle{definition}
\newtheorem{definition}{Definition}[section]

\title{Explainable Prediction of Text Complexity: The Missing Preliminaries for Text Simplification}

\author{Cristina G\^arbacea$^{1}$, Mengtian Guo$^{4}$, Samuel Carton$^{3}$, Qiaozhu Mei$^{1,2}$ \\
$^{1}$Department of EECS, University of Michigan, Ann Arbor \\
$^{2}$School of Information, University of Michigan, Ann Arbor \\
$^{3}$Department of CS, University of Colorado, Boulder \\
$^{4}$School of Information and Library Science, University of North Carolina, Chapel Hill \\
}

\date{}

\begin{document}
\maketitle
\begin{abstract}
  Text simplification reduces the language complexity of professional content for accessibility purposes. 
  End-to-end neural network models have been widely adopted to directly generate the simplified version of input text, usually functioning as a blackbox. We show that text simplification can be decomposed into a compact pipeline of tasks to ensure the transparency and explainability of the process.  The first two steps in this pipeline are often neglected: 1) to predict whether a given piece of text needs to be simplified, and 2) if yes, to identify complex parts of the text.  The two tasks can be solved separately using either lexical or deep learning methods, or solved jointly.  
 Simply applying explainable complexity prediction as a preliminary step, the out-of-sample text simplification performance of the state-of-the-art, black-box simplification models can be improved by a large margin. 
  
\end{abstract}

\section{Introduction}
Text simplification aims to reduce the language complexity of highly specialized textual content so that it is accessible for readers who lack adequate literacy skills, such as children, people with low education, people who have reading disorders or dyslexia, and non-native speakers of the language. 

Mismatch between language complexity and literacy skills is identified as a critical source of bias and inequality in the consumers of systems built upon processing and analyzing professional text content.  Research has found that it requires on average 18 years of education for a reader to properly understand the clinical trial descriptions on ClinicalTrials.gov, and this introduces a potential self-selection bias to those trials \cite{wu2016assessing}.  


Text simplification 
has considerable potential to improve the fairness and transparency of text information systems.  Indeed, the Simple English Wikipedia (\url{simple.wikipedia.org}) has been constructed to disseminate Wikipedia articles to kids and English learners. In healthcare, consumer vocabulary are used to replace professional medical terms to better explain medical concepts to the public \cite{abrahamsson2014medical}. In education, natural language processing and simplified text generation technologies are believed to have the potential to improve student outcomes and bring equal opportunities for learners of all levels in teaching, learning and assessment \cite{mayfield2019equity}.

Ironically, the definition of ``text simplification'' in literature has never been transparent. The term may refer to reducing the complexity of text at various linguistic levels, ranging all the way through replacing individual words in the text to generating a simplified document completely through a computer agent. 
In particular, \textit{lexical simplification} \cite{devlin1999simplifying} is concerned with replacing complex words or phrases with simpler alternatives; \textit{syntactic simplification} \cite{siddharthan2006syntactic} alters the syntactic structure of the sentence; \textit{semantic simplification} \cite{kandula2010semantic} paraphrases portions of the text into simpler and clearer variants. 
More recent approaches simplify texts in an end-to-end fashion, employing machine translation models in a monolingual setting regardless of the type of simplifications \cite{dress, guo2018dynamic, van2019evaluating}.  Nevertheless, these models are limited on the one hand due to the absence of large-scale parallel (complex $\rightarrow$ simple) monolingual training data, and on the other hand 
due to the lack of interpretibility of their black-box procedures \cite{alva2017learning}. 

Given the ambiguity in problem definition, 
there also lacks consensus on how to measure the goodness of text simplification systems, and automatic evaluation measures are perceived ineffective and sometimes detrimental to the specific procedure, in particular when they favor shorter but not necessarily simpler sentences \cite{napoles2011evaluating}. While end-to-end simplification models demonstrate superior performance on benchmark datasets, their success is often compromised in out-of-sample, real-world scenarios \cite{d2020underspecification}. 
Our work is motivated by the aspiration that increasing the transparency and explainability of a machine learning procedure may help its generalization into unseen scenarios \cite{doshi2018considerations}. We show that the general problem of text simplification can be formally decomposed into a compact and transparent pipeline of modular tasks.  We present a systematic analysis of the first two steps in this pipeline, which are commonly overlooked:  1) \textit{ to predict whether a given piece of text needs to be simplified at all}, and 2) \textit{to identify which part of the text needs to be simplified}. 
The second task can also be interpreted as an explanation of the first task: why a piece of text is considered complex. These two tasks can be solved separately, using either lexical or deep learning methods, or they can be solved jointly through an end-to-end, explainable predictor.  Based on the formal definitions, we propose general evaluation metrics for both tasks and empirically compare a diverse portfolio of methods using multiple datasets from different domains, including news, Wikipedia, and scientific papers. We demonstrate that by simply applying explainable complexity prediction as a preliminary step, the out-of-sample text simplification performance of the state-of-the-art, black-box models can be improved by a large margin. 

Our work presents a promising direction towards a transparent and explainable solution to text simplification in various domains. 

\section{Related Work}


\subsection {Text Simplification}

\subsubsection{Identifying complex words} 



Text simplification at word level has been done through 1) \textbf{lexicon based} approaches, which match words to lexicons of complex/simple words 
\cite{deleger2009extracting, elhadad2007mining}, 2) \textbf{threshold based} approaches, which apply a threshold over word lengths or certain statistics \cite{leroy2013user}, 3) \textbf{human driven} approaches, which solicit the user's input on which words need simplification \cite{rello2013simplify}, and 4) \textbf{classification} methods, which train machine learning models to distinguish complex words from simple words \cite{shardlow2013comparison}. Complex word identification is also the main topic of SemEval 2016 Task 11 \cite{paetzold2016semeval}, aiming to determine whether a non-native English speaker can understand the meaning of a word in a given sentence. Significant differences exist between simple and complex words, and the latter on average are shorter, less ambiguous, less frequent, and more technical in nature.  Interestingly, the frequency of a word is identified as a reliable indicator of its simplicity \cite{leroy2013user}. 

While the above techniques have been widely employed for complex word identification, the results reported in the literature are rather controversial and 
it is not clear to what extent one technique outperforms the other in the absence of standardized high quality parallel corpora for text simplification \cite{paetzold2015reliable}. Pre-constructed lexicons are often limited and do not generalize to different domains. It is intriguing that classification methods reported in the literature are not any better than a ``simplify-all'' baseline \cite{shardlow2014survey}.  


\subsubsection{Readability assessment}
Traditionally, measuring the level of reading difficulty is done through lexicon and rule-based metrics  
such as the age of acquisition lexicon (AoA) \cite{kuperman2012age} and the Flesch-Kincaid Grade Level \cite{kincaid1975derivation}.  A machine learning based approach in \cite{schumacher2016predicting} extracts lexical, syntactic, and discourse features and train logistic regression classifiers to predict the relative complexity of a single sentence in a pairwise setting. The most predictive features are simple representations based on AoA norms. The perceived difficulty of a sentence is highly influenced by properties of the surrounding passage. 
Similar methods are used for fine-grained classification of text readability \cite{aluisio2010readability} and complexity \cite{vstajner2020shallow}.

\subsubsection{Computer-assisted paraphrasing} 

Simplification rules are learnt by finding words from a complex sentence that correspond to different words in a simple sentence \cite{alva2017learning}.  Identifying simplification operations such as copies, deletions, and substitutions for words from parallel complex vs. simple corpora helps understand how human experts simplify text \cite{alva2017learning}. Machine translation has been employed to learn phrase-level alignments for sentence simplification \cite{wubben2012sentence}. Lexical and phrasal paraphrase rules are extracted in \cite{pavlick2016simple}. These methods are often evaluated by comparing their output to gold-standard, human-generated simplifications, using standard metrics (e.g., token-level precision, recall, F1), machine translation metrics (e.g., BLEU \cite{papineni2002bleu} 
), text simplification metrics (e.g. SARI \cite{xu2016optimizing} which rewards copying words from the original sentence), and readability metrics (among which Flesch-Kincaid Grade Level \cite{kincaid1975derivation} and Flesch Reading Ease \cite{kincaid1975derivation} are most commonly used). It is desirable that the output of the computational models is ultimately validated by human judges \cite{shardlow2014survey}.




\subsubsection{End-to-end simplification} Neural encoder-decoder models are used to learn simplification rewrites from monolingual corpora of complex and simple sentences \cite{scarton2018learning, van2019evaluating, dress, guo2018dynamic}. 
On one hand, these models often obtain superior performance on particular evaluation metrics, as the neural network directly optimizes these metrics in training. On the other hand, it is hard to interpret what exactly are learned in the hidden layers, and without this transparency it is difficult to adapt these models to new data, constraints, or domains.  For example, these end-to-end simplification models tend not to distinguish whether the input text should or should not be simplified at all, making the whole process less transparent. When the input is already simple, the models tend to oversimplify it and deviate from its original meaning (see Section~\ref{sec:discussion}). 

\subsection{Explanatory Machine Learning}

 Various approaches are proposed in the literature to address the explainability and interpretability of machine learning agents. The task of providing explanations for black-box models has been tackled either at a local level by explaining individual predictions of a classifier \cite{lime}, 
 or at a global level by providing explanations for the model behavior as a whole \cite{letham2015interpretable}. More recently, differential explanations are proposed to describe how the logic of a model varies across different subspaces of interest \cite{lakkaraju2019faithful}. Layer-wise relevance propagation  \cite{arras2017relevant} is used to trace backwards text classification decisions to individual words, which are assigned scores to reflect their separate contribution to the overall prediction. 




LIME \cite{lime} is a model-agnostic explanation technique which can approximate any machine learning model locally with another sparse linear interpretable model. 
SHAP \cite{lundberg2017unified} evaluates Shapley values as the average marginal contribution of a feature value across all possible coalitions by considering all possible combinations of inputs and all possible predictions for an instance. 
Explainable classification can also be solved simultaneously through a neural network, using hard attentions to select individual words into the ``rationale'' behind a classification decision \cite{lei2016rationalizing}. Extractive adversarial networks employs a three-player adversarial game which addresses high recall of the rationale  \cite{carton2018extractive}. The model consists of a generator which extracts an attention mask for each token in the input text, a predictor that cooperates with the generator and makes prediction from the rationale (words attended to), and an adversarial predictor that makes predictions from the remaining words in the inverse rationale. The minimax game between the two predictors and the generator is designed to ensure all predictive signals are included into the rationale. 

No prior work has addressed the explainability of text complexity prediction. 
We fill in this gap. 


\section{An Explainable Pipeline for Text Simplification}
\label{problem_definition}


We propose a unified view of text simplification which is decomposed into several carefully designed sub-problems. These sub-problems 
generalize over many approaches, and they are 
logically dependent on and integratable with one another so that they can be organized into a compact pipeline.  


\vspace{-10pt}
\begin{figure}[htbp]
\centering
  \includegraphics[width=\columnwidth]{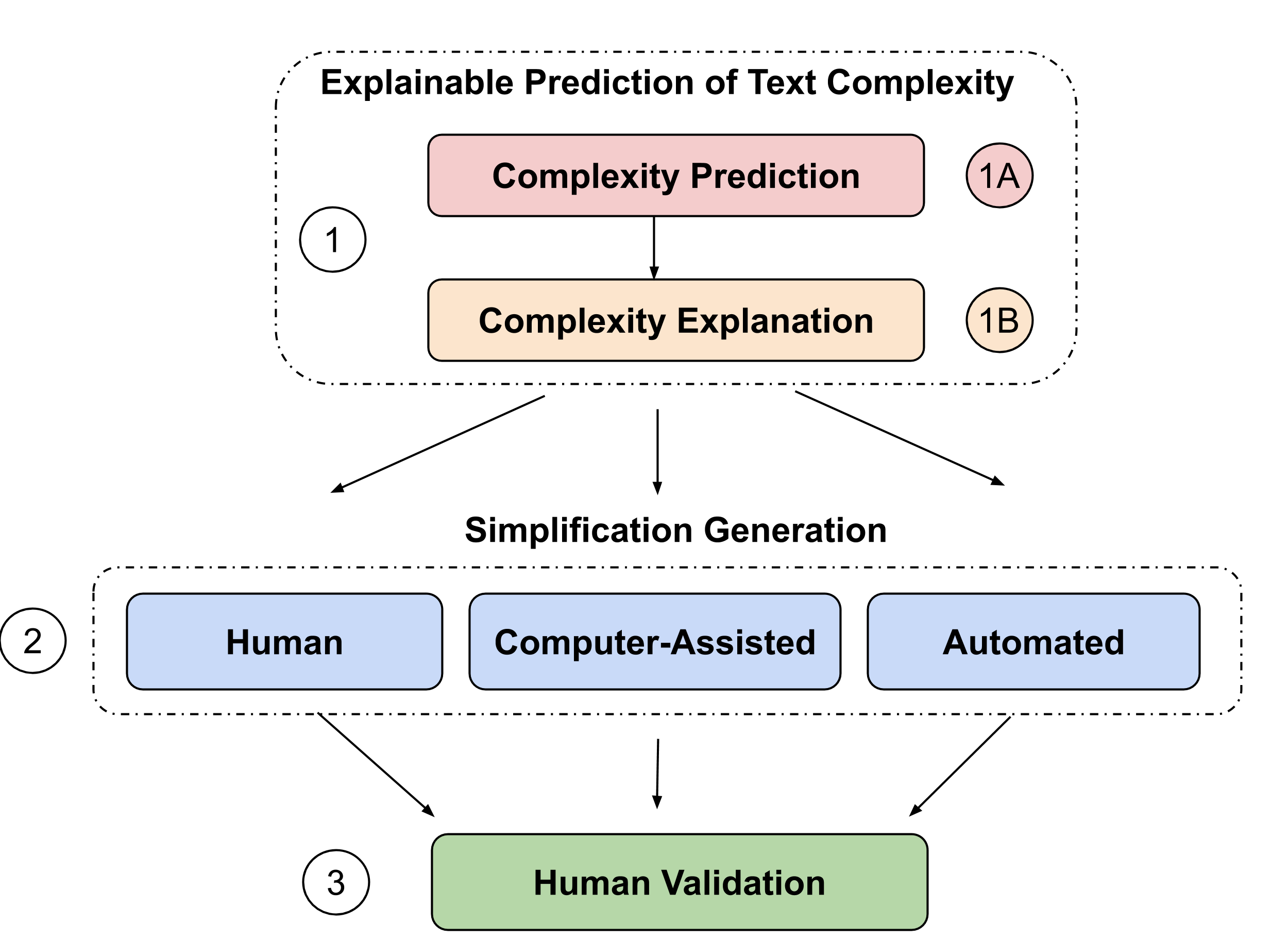}
  \caption{A text simplification pipeline. Explainable prediction of text complexity is the preliminary of any human-based, computer assisted, or automated system. }
  \label{fig::pipeline}
\end{figure}

The first conceptual block in the pipeline (Figure~\ref{fig::pipeline}) is concerned with explainable prediction of the complexity of text. It consists of two sub-tasks: 1) \textit{prediction}: classifying a given piece of text into two categories, needing simplification or not; and 2) \textit{explanation}: highlighting the part of the text that needs to be simplified.  The second conceptual block is concerned with simplification generation, the goal of which is to generate a new, simplified version of the text that needs to be simplified. This step could be achieved through completely manual effort, 
or a computer-assisted approach (e.g., by suggesting alternative words and expressions), or a completely automated method (e.g., by self-translating into a simplified version).  The second building block is piped into a step of human judgment, where the generated simplification is tested, approved, and evaluated by human practitioners.


One could argue that for an automated simplification generation system 
the first block (complexity prediction) is not necessary.  We show that it is not the case.  Indeed, it is unlikely that every piece of text needs to be simplified in reality, and instead the system should first decide whether a sentence needs to be simplified or not.  Unfortunately such a step is often neglected by existing end-to-end simplifiers, 
thus their performance is often biased towards the complex sentences that are selected into their training datasets at the first place and doesn't generalize well to simple inputs. Empirically, when these models are applied to out-of-sample text which shouldn't be simplified at all, they tend to oversimplify the input and result in a deviation from its original meaning (see Section~\ref{sec:discussion}). 

One could also argue that an explanation component (1B) is not mandatory in certain text simplification practices, in particular in an end-to-end neural generative model that does not explicitly identify the complex parts of the input sentence.  In reality, however, it is often necessary to highlight the differences between the original sentence and the simplified sentence (which is essentially a variation of 1B) to facilitate the validation and evaluation of these black-boxes.  More generally, the explainability/interpretability of a machine learning model has been widely believed to be an indispensable factor to its fidelity and fairness when applied to the real world \cite{lakkaraju2019faithful}.  Since the major motivation of text simplification is to improve the fairness and transparency of text information systems, it is critical to explain the rationale behind the simplification decisions, even if they are made through a black-box model. 

Without loss of generality, we can formally define the sub-tasks 1A, 1B, and 2- in the pipeline:

\begin{definition}
(Complexity Prediction). Let text $d \in D$ be a sequence of tokens $w_1w_2...w_n$. The task of complexity prediction is to find a function $f: D \rightarrow \{0, 1\}$ such that $f(d) = 1$ if $d$ needs to be simplified, and $f(d) = 0$ otherwise. 
\end{definition}

\begin{definition}
(Complexity Explanation). Let $d$ be a sequence of tokens $w_1w_2...w_n$ and $f(d) = 1$. The task of complexity explanation/highlighting is to find a function $h: D \rightarrow \{0, 1\}^{n}$ s.t. $h(d) = c_1c_2...c_n$, where $c_i = 1$ means $w_i$ will be highlighted as a complex portion of $d$ and $c_i = 0$ otherwise. We denote $d|h(d)$ as the highlighted part of $d$ and $d|\neg h(d)$ as the unhighlighted part of d. 
\end{definition}

\begin{definition}
(Simplification Generation). Let $d$ be a sequence of tokens $w_1w_2...w_n$ and $f(d) = 1$. The task of simplification generation is to find a function $g: D \rightarrow D'$ s.t. $g(d, f(d), h(d)) = d'$, where $d' = w'_1w'_2...w'_m$ and $f(d') = 0$, subject to the constraint that $d'$ preserves the meaning of $d$.
\end{definition}

In this paper, we focus on an empirical analysis of the first two sub-tasks of explainable prediction of text complexity (1A and 1B), which are the preliminaries of any reasonable text simplification practice.  We leave aside the detailed analysis of simplification generation (2-) for now, as there are many viable designs of $g(\cdot)$ in practice, spanning the spectrum between completely manual and completely automated.  
Since this step is not the focus of this paper, we intend to leave the definition of simplification generation highly general. 

Note that the definitions of complexity prediction and complexity explanation can be naturally extended to a continuous output, where $f(\cdot)$ predicts the complexity level of $d$ and $h(\cdot)$ predicts the complexity weight of $w_i$. The continuous output would align the problem more closely to readability measures \cite{kincaid1975derivation}. In this paper, we stick to the binary output because a binary action (to simplify or not) is almost always necessary in reality even if a numerical score is available. 


Note that the definition of complexity explanation is general enough for existing approaches. In lexical simplification where certain words in a complex vocabulary $V$ are identified to explain the complexity of a sentence, it is equivalent to highlighting every appearance of these words in $d$, or $\forall w_i \in V, c_i = 1$. In automated simplification where there is a self-translation function $g(d) = d'$, $h(d)$ can be simply instantiated as a function that returns a sequence alignment of $d$ and $d'$.  Such reformulation helps us define unified evaluation metrics for complexity explanation (see Section \ref{empirical_analysis}). 

It is also important to note that the dependency between the components, especially complexity prediction and explanation, does not restrict them to be done in isolation. These sub-tasks can be done either separately, or jointly with an end-to-end approach as long as the outputs of $f, h, g$ are all obtained (so that transparency and explainability are preserved). In Section \ref{empirical_analysis}, we include both separate models and end-to-end models for explanatory complexity predication in one shot. 



\section{Empirical Analysis of Complexity Prediction and Explanation}
\label{empirical_analysis}

With the pipeline formulation, we are able to compare a wide range of methods and metrics for the sub-tasks of text simplification.  
We aim to understand how difficult they are in real-world settings and which method performs the best for which task. 

\subsection{Complexity Prediction}
\label{clf_models}

\subsubsection{Candidate Models}

We examine a wide portfolio of deep and shallow binary classifiers to distinguish complex sentences from simple ones. 
Among the shallow models we use Naive Bayes (NB), Logistic Regression (LR), Support Vector Machines (SVM) and Random Forests (RF) 
classifiers trained with unigrams, bigrams and trigrams as features. We also train the classifiers using the lexical and syntactic features proposed in \cite{schumacher2016predicting} combined with the $n$-gram features (denoted as ``enriched features''). We include neural network models such as word and char-level Long Short-Term Memory Network (LSTM) and Convolutional Neural Networks (CNN). We also employ a set of state-of-the-art pre-trained neural language models, 
fine-tuned for complexity prediction; we introduce them below. 

ULMFiT \cite{howard2018universal} 
a language model on a large general corpus such as WikiText-103 
and then fine-tunes it on the target task using 
slanted triangular rates, 
and gradual unfreezing. 
We use the publicly available implementation\footnote{ \url{https://docs.fast.ai/tutorial.text.html}, retrieved on 5/31/2021.} of the model with two fine-tuning epochs for each dataset and the model quickly adapts to a new task.

BERT \cite{devlin2019bert} trains deep bidirectional language representations 
and has greatly advanced the state-of-the-art for many natural language processing tasks.  
The model is pre-trained on the English Wikipedia as well as the Google Book Corpus
. Due to computational constraints, we use the 12 layer BERT base pre-trained model and fine-tune it on our three datasets. We select the best hyperparameters based on each validation set.  


XLNeT \cite{yang2019xlnet} overcomes the limitations of BERT (mainly the use of masks) with a permutation-based objective which considers bidirectional contextual information from all positions without data corruption. 
We use the 12 layer XLNeT base pre-trained model on the English Wikipedia, the Books corpus (similar to BERT), Giga5, 
ClueWeb 2012-B, 
and Common Crawl. 

\subsubsection{Evaluation Metric}

We evaluate the performance of complexity prediction models using \textit{classification accuracy} on balanced training, validation, and testing datasets. 

\subsection{Complexity Explanation}
\label{expl_models}

\subsubsection{Candidate Models}

We use \textit{LIME} in combination with LR and LSTM classifiers, \textit{SHAP} on top of LR, and the \textit{extractive adversarial networks} which jointly conducts complexity prediction and explanation. We feed each test complex sentence as input to these explanatory models and compare their performance at identifying tokens (words and punctuation) that need to be removed or replaced from the input sentence. 


We compare these explanatory models with three baseline methods:
1) \textit{Random highlighting}: randomly draw the size and the positions of tokens to highlight;
	2) \textit{Lexicon based highlighting}: highlight words that appear in the Age-of-Acquisition (AoA) lexicon \cite{kuperman2012age}, which contains ratings for 30,121 English content words (nouns, verbs, and adjectives) indicating the age at which a word is acquired; and 
    3) \textit{Feature highlighting}: highlight the most important features of the best performing LR models for complexity prediction. 

\subsubsection{Evaluation Metrics}
\label{expl_evaluation}

Evaluation of explanatory machine learning is an open problem. In the context of complexity explanation, 
when the ground truth of highlighted tokens ($y_c(d) = c_1c_2...c_n$, $c_i \in \{0, 1\}$) in each complex sentence $d$ is available, we can compare the output of complexity explanation $h(d)$ with $y_c(d)$.  
Such per-token annotations are usually not available in scale. To overcome this, 
given a complex sentence $d$ and its simplified version $d'$, we assume that all tokens $w_i$ in $d$ which are absent in $d'$ are candidate words for deletion or substitution during the text simplification process and should therefore be highlighted in complexity explanation (i.e., $c_i = 1$). 

In particular, we use the following evaluation metrics for complexity explanation: 1) 
\textit{Tokenwise Precision (P)}, which measures the proportion of highlighted tokens in $d$ that are truly removed in $d'$;  
2) \textit{Tokenwise Recall (R)}, which measures the proportion of tokens removed in $d'$ that are actually highlighted in $d$;    
3) \textit{Tokenwise F1}, the harmonic mean of $P$ and $R$; 
4) word-level \textit{Edit distance (ED)} \cite{levenshtein1966binary}: 
between the \textbf{unhighlighted} part of $d$ and the simplified document $d'$. Intuitively, a more successful complexity explanation would highlight most of the tokens that need to be simplified, thus the remaining parts in the complex sentences will be closer to the simplified version, achieving a lower edit distance (we also explore ED with a higher penalty cost for the substitution operation, namely values of 1, 1.5 and 2); and  
5) \textit{Translation Edit Rate (TER)} \cite{snover2006study},
which measures the minimum number of edits needed to change a hypothesis (the unhighlighted part of $d$) so that it exactly matches the closest references (the simplified document $d'$). 
Note these metrics are all proxies of the real editing process from $d$ to $d'$. When token-level edit history is available (e.g., through track changes), it is better to compare the highlighted evaluation with these true changes made. 
We compute all the metrics at sentence level and macro-average them. 

\subsection{Experiment Setup}

\subsubsection{Datasets}

We use three different datasets (Table \ref{table_dataset}) which cover different domains and application scenarios of text simplification. Our first dataset is \textit{Newsela} \cite{xu2015problems}, a corpus of news articles simplified by professional news editors. 
In our experiments we use the parallel Newsela corpus with the training, validation, and test splits made available in \cite{dress}.  Second, we use the \textit{WikiLarge} corpus introduced in \cite{dress}. The training subset of WikiLarge is created by assembling datasets of parallel aligned Wikipedia - Simple Wikipedia sentence pairs available in the literature \cite{kauchak2013improving}.
While this training set is obtained through automatic alignment procedures which can be noisy, the validation and test subsets of WikiLarge contain complex sentences with simplifications provided by Amazon Mechanical Turk workers \cite{xu2016optimizing}; we increase the size of validation and
test on top of the splits made available in \cite{dress}. 
Third, we use the dataset released by the \textit{Biendata} competition\footnote{\url{https://www.biendata.com/competition/hackathon}, retrieved on 5/31/2021.}, which asks participants to match research papers from various scientific disciplines with press releases that describe them. Arguably, rewriting scientific papers into press releases has mixed objectives that are not simply text simplification. We include this task to test the generalizability of our explainable pipeline (over various definitions of simplification).  
We use alignments at title level.  
On average, a complex sentence in Newsela, WikiLarge, Biendata contains 23.07, 25.14, 13.43 tokens, and the corresponding simplified version is shorter, with 12.75, 18.56, 10.10 tokens.

\begin{table}[!h]
\caption{Aligned complex-simple sentence pairs. } 
\centering
\scalebox{0.8}{
\begin{tabular}{l|  l | l | l  }
\hline
\hline
\textbf{Dataset} & \textbf{Training} & \textbf{Validation} & \textbf{Test} \\
\hline
\hline
\textit{Newsela} & 94,208 pairs & 1,129 pairs & 1,077 pairs \\
\hline
\textit{WikiLarge} & 208,384 pairs & 29,760 pairs & 59,546 pairs\\
\hline
\textit{Biendata} & 29,700 pairs & 4,242 pairs & 8,486 pairs \\
\hline
\hline
\end{tabular}}
\label{table_dataset}
\end{table}

\subsubsection{Ground Truth Labels}

The original datasets contain aligned complex-simple sentence pairs instead of classification labels for complexity prediction. We infer ground-truth complexity labels for each sentence such that: \textit{label 1} is assigned to every sentence for which there is an aligned simpler version not identical to itself (the sentence is complex and needs to be simplified); \textit{label 0} is assigned to all simple counterparts of complex sentences, as well as to those sentences that have corresponding ``simple'' versions identical to themselves (i.e., these sentences do not need to be simplified). For complex sentences that have label 1, we further identify which tokens are not present in corresponding simple versions. 




\subsubsection{Model Training}


For all shallow and deep classifiers we find the best hyperparameters using random search 
on validation, with early stopping. We use grid search on validation to fine-tune hyperparameters of the pre-trained models, such as maximum sequence length, batch size, learning rate, and number of epochs. For ULMFit on Newsela, we set batch size to 128 and learning rate to 1e-3. For BERT on WikiLarge, batch size is 32, learning rate is 2e-5, and maximum sequence length is 128. For XLNeT on Biendata, batch size is 32, learning rate is 2e-5, and maximum sequence length is 32.

We use grid search on validation to fine-tune the complexity explanation models, including the extractive adversarial network. For LR and LIME we determine the maximum number of words to highlight based on TER score on validation (please see Table \ref{table_LIME_hyperparam}); 
for SHAP we highlight all features with positive assigned weights, all based on TER. 
\begin{table}[!htbp]
\caption{Maximum numbers of most important LR features and features highlighted by LIME. } 
\centering
\scalebox{0.73}{
\begin{tabular}{l|  l | l | l  }
\hline
\hline
\textbf{Model} & \textbf{Newsela} & \textbf{WikiLarge} & \textbf{Biendata} \\
\hline
\hline
\textit{LR} & 200 features & 20,000 features & 200 features \\
\hline
\textit{LIME \& LR} & 10 features & 50 features & 10 features \\
\hline
\textit{LIME \& LSTM} & 60 features & 20 features & 40 features \\
\hline
\hline
\end{tabular}}
\label{table_LIME_hyperparam}
\end{table}

For extractive adversarial networks 
batch size is set to 256, learning rate is 1e-4, and adversarial weight loss equals 1; in addition, sparsity weight is 1 for Newsela and Biendata, and 0.6 for WikiLarge; lastly, coherence weight is 0.05 for Newsela, 0.012 for WikiLarge, and 0.0001 for Biendata. 
\vspace{-5pt}
\section{Results}

\subsection{Complexity Prediction}
In Table \ref{table_results_classification}, we evaluate how well the representative shallow, deep, and pre-trained classification models can determine whether a sentence needs to be simplified at all. We test for statistical significance of the best classification results compared to all other models using a two-tailed z-test. 

\begin{table}[!h]
\caption{Accuracy of representative shallow$^*$, deep, and pre-trained models for complexity prediction. \textbf{BOLD}: best performing models. } 
\scalebox{0.65}{
\begin{tabular}{l|  c | c | c  }
\hline
\hline
\textbf{Classifier} & \textbf{Newsela} & \textbf{WikiLarge} & \textbf{Biendata} \\
\hline
\textit{NB n-grams} & 73.10 \% & 62.70 \% & 84.30 \% \\
\hline
\textit{NB enriched features} & 73.10 \% & 63.10 \% & 86.00 \% \\
\hline
\textit{LR n-grams} & 75.30 \% & 71.90 \% & 89.60 \%\\
\hline
\textit{LR enriched features } & 76.30 \% & 72.60 \% & 91.70 \% \\
\hline
\textit{SVM n-grams} & 75.20 \% & 71.90 \% & 89.50 \% \\
\hline
\textit{SVM enriched features} & 77.39 \% & 70.16 \% & 88.60 \% \\
\hline
\textit{RF n-grams} & 71.50 \% & 71.50 \% & 84.60 \% \\
\hline
\textit{RF enriched features} & 74.40 \% & 73.40 \% & 87.00 \% \\
\hline
\hline
\textit{LSTM (word-level)} & 73.31 \%  & 71.62 \% & 89.87 \%\\
\hline
\textit{CNN (word-level)}  & 70.71 \% & 69.27 \% & 89.05 \% \\
\hline
\textit{CNN (char-level)}  & $78.83 \% ^{\dagger}$ & 74.88 \% & 88.00 \%\\
\hline
\textit{CNN (word \& char-level)}  & 75.90 \% & 74.00 \% & 92.30 \% \\
\hline
\textit{Extractive Adversarial Networks}  & 72.76 \% & 71.50 \% & 88.64 \% \\
\hline
\hline
\textit{ULMFiT} & $\textbf{80.83\%} ^{**}$ & 74.80 \% & 94.17 \%\\
\hline
\textit{BERT} & 77.15 \% & $\textbf{81.45\%} ^{**}$  & 94.43 \%\\
\hline
\textit{XLNeT} & $78.83 \%^{\dagger} $ & 73.49 \% & $\textbf{95.48\%} ^{**}$ \\
\hline
\hline
\end{tabular}}
\scriptsize * Shallow models perform similarly and some are omitted for space; Difference between the best performing model and other models is statistically significant: $p < 0.05$ (*), $p < 0.01$ (**), except for $\dagger$: difference between this model and the best performing model is not statistically significant.  
\label{table_results_classification}
\end{table}

In general, the best performing models can achieve around 80\% accuracy on two datasets (Newsela and WikiLarge) and a very high performance on the Biendata ($>95\%$). This difference presents the difficulty of complexity prediction in different domains, and distinguishing highly specialized scientific content from public facing press releases is relatively easy (Biendata). 

Deep classification models in general outperform shallow ones, however with carefully designed handcrafted features and proper hyperparameter optimization shallow models tend to approach to the results of the deep classifiers.  Overall models pre-trained on large datasets and fine-tuned for text simplification yield superior classification performance. For Newsela the best performing classification model is ULMFiT (accuracy = 80.83\%, recall = 76.87\%), which significantly (p $<$ 0.01) surpasses all other classifiers except for XLNeT and CNN (char-level).  On WikiLarge, BERT presents the highest accuracy ($81.45\%, p < 0.01$), and recall = 83.30\%.  On Biendata, XLNeT yields the highest accuracy ($95.48\%, p < 0.01$) with recall = 94.93\%, although the numerical difference to other pre-trained language models is small. This is consistent with recent findings in other natural language processing tasks \cite{cohan2019pretrained}. 

\subsection{Complexity Explanation}

We evaluate how well complexity classification can be explained, or how accurately the complex parts of a sentence can be highlighted.

\begin{table}[!h]
\caption{Results for complexity explanation. P, R and F1 - the higher the better; TER and ED 1.5 - the lower the better. \textbf{BOLD} \& \underline{Underlined}: best \& second best. }
\centering
\scalebox{0.62}{
\begin{tabular}{l| l | c|  c | c | c | c }
\hline
\hline
\textbf{Dataset} & \textbf{Explanation Model} & \textbf{P} & \textbf{R} & \textbf{F1} & \textbf{TER} &  \textbf{ED 1.5} \\
\hline
\hline
\multirow{7}{*}{Newsela} & Random & 0.515 & 0.487 & 0.439 & 0.985 & 13.825 \\
\cline{2-7}
& AoA lexicon & \textbf{0.556} & 0.550 & 0.520 & 0.867 & 12.899 \\
\cline{2-7}
& LR Features & 0.522 & 0.250 & 0.321 & 0.871 & 12.103 \\
\cline{2-7}
& LIME \& LR  & 0.535 & 0.285 & 0.343 & 0.924 & 12.459 \\
\cline{2-7}
& LIME \& LSTM  & 0.543 & \textbf{0.818} & \textbf{0.621} & 0.852 & \underline{11.991} \\
\cline{2-7}
& SHAP \& LR  & \underline{0.553} & \underline{0.604} & \underline{0.546} & \underline{0.848} & 12.656 \\
\cline{2-7}
& Extractive Networks & 0.530 & 0.567 & 0.518 & \textbf{0.781} & \textbf{11.406}  \\
\hline \hline
\multirow{7}{*}{WikiLarge} & Random & 0.412 & 0.439 & 0.341 & 1.546 & 17.028 \\
\cline{2-7}
 & AoA lexicon & 0.427 & 0.409 & 0.357 & 1.516 & \underline{16.731} \\
 \cline{2-7}
& LR Features & 0.442 & \underline{0.525} & 0.413 & \underline{0.993} & 17.933  \\
\cline{2-7}
& LIME \& LR  & 0.461 & 0.509 & 0.415 & \textbf{0.988} & 18.162 \\
\cline{2-7}
& LIME \& LSTM  & \textbf{0.880} & 0.470 & \underline{0.595} & 1.961 & 25.051 \\
\cline{2-7}
& SHAP \& LR  & \underline{0.842} & \textbf{0.531} & \textbf{0.633} & 1.693 & 22.811 \\
\cline{2-7}
& Extractive Networks  & 0.452 & 0.429 & 0.359 & 1.434 & \textbf{16.407} \\
\hline \hline
\multirow{7}{*}{Biendata} & Random & 0.743 & 0.436 & 0.504 & 1.065 & 12.921 \\
\cline{2-7}
 & AoA lexicon & 0.763 & 0.383 & 0.475 & 1.064 & 13.247 \\
 \cline{2-7}
& LR Features & 0.796 & 0.257 & 0.374 & 0.979 & 10.851 \\
\cline{2-7}
& LIME \& LR  & \textbf{0.837} & 0.466 & 0.577 & 0.982 & \textbf{10.397} \\
\cline{2-7}
& LIME \& LSTM  & \underline{0.828} & \underline{0.657} & \underline{0.713} & \textbf{0.952} & 16.568 \\
\cline{2-7}
& SHAP \& LR  & 0.825 & 0.561 & 0.647 & 0.979 & 11.908 \\
\cline{2-7}
& Extractive Networks & 0.784 & \textbf{0.773} & \textbf{0.758} & \underline{0.972} & \underline{10.678}  \\
\hline
\hline
\end{tabular}}
\label{table_complexity_explanation}
\end{table}

Results (Table~\ref{table_complexity_explanation}) show that highlighting words in the AoA lexicon or LR features are rather strong baselines, indicating that most complexity of a sentence still comes from word usage.  Highlighting more LR features leads to a slight drop in precision and a better recall. Although LSTM and LR perform comparably on complexity classification, using LIME to explain LSTM presents better recall, F1, and TER (at similar precision) compared to using LIME to explain LR.  
The LIME \& LSTM combination is reasonably strong on all datasets, as is SHAP \& LR. 
TER is a reliable indicator of the difficulty of the remainder (unhighlighted part) of the complex sentence. ED with a substitution penalty of 1.5 efficiently captures the variations among the explanations.  On Newsela and Biendata, the extractive adversarial networks yield solid performances (especially TER and ED 1.5), indicating that jointly making predictions and generating explanations reinforces each other. 
Table \ref{table_explanations} provides examples of highlighted complex sentences by each explanatory model.

\begin{table*}[!htbp]
\caption{Explanations of complexity predictions (in red). Extractive network obtains a higher recall. } 
\centering
\small
\scalebox{1}{
\begin{tabular}{l| l }
\hline
\hline
\textbf{Explanatory Model} & \textbf{ Complexity Explanation} \\
\hline
LIME \& LR & \textcolor{red}{Their} fatigue changes \textcolor{red}{their} voices , \textcolor{red}{but} they 're still \textcolor{red}{on the} freedom highway .\\
\hline
LIME \& LSTM & \textcolor{red}{Their fatigue changes their voices , but they 're still on the freedom highway} .\\
\hline
SHAP \& LR & \textcolor{red}{Their fatigue} changes \textcolor{red}{their} voices , \textcolor{red}{but} they 're \textcolor{red}{still on the freedom highway} . \\
\hline
Extractive Networks & \textcolor{red}{Their fatigue changes their voices ,} but they \textcolor{red}{'re} still on the \textcolor{red}{freedom} highway .\\
\hline
Simple sentence & Still , they are fighting for their rights .\\
\hline
\hline
LIME \& LR & Digitizing \textcolor{red}{physically} preserves these \textcolor{red}{fragile} papers \textcolor{red}{and} allows people \textcolor{red}{to} see them , he \textcolor{red}{said} . \\
\hline
LIME \& LSTM & Digitizing \textcolor{red}{physically} preserves \textcolor{red}{these fragile papers and allows people to see them , he said} .\\
\hline
SHAP \& LR & Digitizing \textcolor{red}{physically} preserves \textcolor{red}{these} fragile papers \textcolor{red}{and} allows people \textcolor{red}{to see them} , he \textcolor{red}{said} . \\
\hline
Extractive Networks & \textcolor{red}{Digitizing physically preserves} these \textcolor{red}{fragile} papers \textcolor{red}{and allows} people to see them , he said .\\
\hline
Simple sentence & The papers are old and fragile , he said .\\
\hline
\hline
\end{tabular}}
\label{table_explanations}
\end{table*}

\subsection{Benefit of Complexity Prediction}
\label{sec:discussion}
One may question whether explainable prediction of text complexity is still a necessary preliminary step in the pipeline if a strong, end-to-end simplification generator is used. We show that it is. 
We consider the scenario where a pre-trained, end-to-end text simplification model is blindly applied to texts regardless of their complexity level, compared to only simplifying those considered complex by the best performing complexity predictor in Table~\ref{table_results_classification}. Such a comparison demonstrates whether adding complexity prediction as a preliminary step is beneficial to a text simplification process when a state-of-the-art, end-to-end simplifier is already in place. 
From literature we select the current best text simplification models on WikiLarge and Newsela which have released pre-trained models: 

\begin{itemize}
    \item ACCESS \cite{martin2019controllable}, a controllable sequence-to-sequence simplification model that reported the  
    highest performance (41.87 SARI) on WikiLarge.  
    
    \item Dynamic Multi-Level Multi-Task Learning for Sentence Simplification (DMLMTL) \cite{guo2018dynamic}, 
    which reported the highest performance (33.22 SARI) on Newsela.
    
\end{itemize}

We apply the author-released, pre-trained ACCESS and DMLMTL on all sentences from the validation and testing sets of all three datasets. We do not use the training examples as the pre-trained models may have already seen them. Presumably, a smart model should \textbf{not} further simplify an input sentence if it is already simple enough.  However, to our surprise, a majority of the \textit{out-of-sample simple} sentences are still changed by both models (above 90\% by DMLMTL and above 70\% by ACCESS, please see Table \ref{table_simplifications}). 

\begin{table}[!h]
\caption{Percentage of out-of-sample simple sentences changed by pre-trained, end-to-end simplification models. Ideal value is 0\%. }
\centering
\scalebox{0.8}{
\begin{tabular}{l| l | c|  c }
\hline
\hline
\textbf{Dataset} & \textbf{ Pre-trained Model} & \textbf{Validation} & \textbf{Testing} \\
\hline
\hline
\multirow{2}{*}{Newsela} & ACCESS & 72.73 \% & 75.50 \%  \\
\cline{2-4}
& DMLMTL & 90.48 \% & 91.69 \% \\
\hline \hline
\multirow{2}{*}{WikiLarge} & ACCESS &  70.83 \% & 71.12 \% \\
\cline{2-4}
& DMLMTL & 95.20 \% & 95.61 \% \\
\hline \hline
\multirow{2}{*}{Biendata} & ACCESS & 94.25 \% & 93.66 \% \\
\cline{2-4}
& DMLMTL & 98.88 \% & 98.73 \% \\
\hline
\hline
\end{tabular}}
\label{table_simplifications}
\end{table}

We further quantify the difference with vs. without complexity prediction as a preliminary step. Intuitively, without complexity prediction, an already simple sentence is likely to be overly simplified and result in a loss in text simplification metrics. In contrast, an imperfect complexity predictor may mistaken a complex sentence as simple, which misses the opportunity of simplification and results in a loss as well. The empirical question is which loss is higher. From Table~\ref{table_simplifications_ACCESS}, we see that after directly adding a complexity prediction step before either of the state-of-the-art simplification models, there is a considerable drop of errors in three text simplification metrics: Edit Distance (ED), TER, and Fr\'echet Embedding Distance (FED) that measures the difference of a simplified text and the ground-truth in a semantic space \cite{de2019training}. For ED alone, the improvements are between 30\% to 50\%. This result is very encouraging: considering that the complexity predictors are only 80\% accurate and the complexity predictor and the simplification models don't depend on each other, there is considerable room to optimize this gain. Indeed, the benefit is higher on Biendata where the complexity predictor is more accurate.

\begin{table}[!h]
\caption{
Out-of-sample performance of simplification models. 
ED, TER, FED metrics: the lower the better. 
Adding complexity prediction as preliminary step reduces simplification error by a wide margin. 
}
\centering
\scalebox{0.55}{
\begin{tabular}{l| l | l | c|  c }
\hline
\hline
\textbf{Dataset} & \textbf{Sentence Pairs} & \textbf{Metric} &  \textbf{ACCESS} &\textbf{DMLMTL} \\
\hline
\hline
\multirow{9}{*}{Newsela} & \multirow{3}{*}{No complexity prediction } & ED & 4.044 & 12.212 \\
\cline{3-5}
& \multirow{3}{*}{(simplify everything) } & TER & 0.175 & 1.611 \\
\cline{3-5}
& & FED & 0.016 & 0.170 \\
\cline{2-5}
& \multirow{3}{*}{With complexity prediction} & ED & \textbf{2.631} (-35\%) & \textbf{8.677} (-29\%) \\
\cline{3-5}
& \multirow{3}{*}{(predicted simple: no change)} & TER & 0.089 (-49\%) & 1.149 (-29\%) \\
\cline{3-5}
& & FED & 0.006 (-63\%) & 0.066 (-61\%) \\

\hline \hline
\multirow{9}{*}{WikiLarge} & \multirow{3}{*}{No Complexity Prediction} & ED & 5.857 & 16.920 \\
\cline{3-5}
& \multirow{3}{*}{(simplify everything) } & TER & 0.208 & 2.328 \\
\cline{3-5}
& & FED & 0.004 & 0.143 \\
\cline{2-5}
& \multirow{3}{*}{With Complexity Prediction}& ED & \textbf{4.021} (-31\%) & \textbf{10.566} (-38\%) \\
\cline{3-5}
& \multirow{3}{*}{(predicted simple: no change)} & TER & 0.132 (-37\%) & 1.452 (-38\%) \\
\cline{3-5}
& & FED & 0.002 (-50\%) & 0.049 (-66\%) \\ 

\hline \hline
\multirow{9}{*}{Biendata} & \multirow{3}{*}{No Complexity Prediction} & ED & 3.796 & 9.030 \\
\cline{3-5}
& \multirow{3}{*}{(simplify everything) } & TER & 0.254 & 1.348 \\
\cline{3-5}
& & FED & 0.033 & 0.131 \\
\cline{2-5}
& \multirow{3}{*}{With Complexity Prediction} & ED & \textbf{1.887} (-50\%) & \textbf{5.249} (-42\%) \\
\cline{3-5}
& \multirow{3}{*}{(predicted simple: no change)} & TER & 0.114 (-55\%) & 0.819 (-39\%) \\
\cline{3-5}
& & FED & 0.009 (-73\%) & 0.051 (-61\%) \\ 
\hline
\hline
\end{tabular}}
\label{table_simplifications_ACCESS}
\end{table}

Qualitatively, one could frequently observe syntactic, semantic, and logical mistakes in the model-simplified version of \textit{simple} sentences. We give a few examples below.  

\begin{itemize}
    \item  In Ethiopia, HIV disclosure is low $\rightarrow$ In Ethiopia , HIV is low (ACCESS)
    \item   Mustafa Shahbaz , 26 , was shopping for books about science . $\rightarrow$ Mustafa Shahbaz , 26 years old , was a group of books about science . (ACCESS)
    \item New biomarkers for the diagnosis of Alzheimer’s $\rightarrow$ New biomarkers are diagnosed with Alzheimer (ACCESS)
   \item  Healthy diet linked to lower risk of chronic lung disease $\rightarrow$ Healthy diet linked to lung disease (DMLMTL)
    
   \item   Dramatic changes needed in farming practices to keep pace with climate change $\rightarrow$ changes needed to cause climate change (DMLMTL)


   \item  Social workers can help patients recover from mild traumatic brain injuries $\rightarrow$ Social workers can cause better problems . (DMLMTL)
\end{itemize}

All these qualitative and quantitative results suggest that the state-of-the-art black-box models tend to oversimplify and distort the meanings of out-of-sample input that is already simple. Evidently, the lack of transparency and explainability has limited the application of these end-to-end black-box models in reality, especially to out-of-sample data, context, and domains. The pitfall can be avoided with the proposed pipeline and simply with explainable complexity prediction as a preliminary step. Even though this explainable preliminary does not necessarily reflect how a black-box simplification model ``thinks'', adding it to the model is able to yield better out-of-sample performance.  


 \section{Conclusions}
 
  We formally decompose the ambiguous notion of text simplification into a compact, transparent, and logically dependent pipeline of sub-tasks, where explainable prediction of text complexity is identified as the preliminary step. 
  We conduct a systematic analysis of its two sub-tasks, namely complexity prediction and complexity explanation, and show that they can be either solved separately or jointly through an extractive adversarial network.  
  While pre-trained neural language models achieve significantly better performance on complexity prediction, an extractive adversarial network that solves the two tasks jointly presents promising advantage in complexity explanation.  Using complexity prediction as a preliminary step reduces the error of the state-of-the-art text simplification models by a large margin. Future work should integrate rationale extractor into the pre-trained neural language models and extend it for simplification generation.  

\section*{Acknowledgement}

This work is in part supported by the National Science Foundation under grant numbers 1633370 and 1620319 and by the National Library of Medicine under grant number 2R01LM010681-05.

\bibliographystyle{acl_natbib}
\bibliography{anthology,acl2021}


\end{document}